\documentclass[10pt,twocolumn,letterpaper]{article}

\usepackage{iccv}
\usepackage{times}
\usepackage{epsfig}
\usepackage{graphicx}
\usepackage{amsmath}
\usepackage{amssymb}

\usepackage{multirow}
\usepackage{booktabs}
\usepackage{arydshln}

\usepackage{caption}
\usepackage{subcaption}

\usepackage[pagebackref=true,breaklinks=true,letterpaper=true,colorlinks,bookmarks=false]{hyperref}

\iccvfinalcopy %

\def\iccvPaperID{1817} %

\ificcvfinal\fi

\begin{document}

\title{ViewFormer: View Set Attention for Multi-view 3D Shape Understanding}

\author{Hongyu Sun\quad Yongcai Wang\quad Peng Wang \quad Xudong Cai \quad Deying Li\\
Renmin University of China\\
No. 59, Zhongguancun Street, Haidian District, Beijing 100872, China\\
{\tt\small \{sunhongyu,ycw,peng.wang,xudongcai,deyingli\}@ruc.edu.cn}
}

\maketitle
\ificcvfinal\thispagestyle{empty}\fi

\begin{abstract}
   This paper presents \emph{ViewFormer}, a simple yet effective model for multi-view 3d shape recognition and retrieval. 
   We systematically investigate the existing methods for aggregating multi-view information and propose a novel ``view set" perspective, which minimizes the relation assumption about the views and releases the representation flexibility. 
   We devise an adaptive attention model to capture pairwise and higher-order correlations of the elements in the view set. 
   The learned multi-view correlations are aggregated into an expressive view set descriptor for recognition and retrieval. 
   Experiments show the proposed method unleashes surprising capabilities across different tasks and datasets. 
   For instance, with only 2 attention blocks and 4.8M learnable parameters, ViewFormer reaches 98.8\% recognition accuracy on ModelNet40 for the first time, exceeding previous best method by 1.1\% . 
   On the challenging RGBD dataset, our method achieves 98.4\% recognition accuracy, which is a 4.1\% absolute improvement over the strongest baseline. 
   ViewFormer also sets new records in several evaluation dimensions of 3D shape retrieval defined on the SHREC'17 benchmark.
\end{abstract}

\section{Introduction}

With the advancement of 3D perception devices and methods, 3D assets (point clouds, meshes, RGBD images, CAD models, \etc) 
become more and more common in daily life and industrial production. 
3D object recognition and retrieval are basic requirements for understanding the 3D contents and the development of these technologies will benefit downstream applications like VR/AR/MR, 3D printing,  and autopilot. 
Existing methods for 3D shape analysis can be roughly divided into three categories according to the input representation: 
point-based~\cite{qi17pointnet,qi17pointnet2,wang19dgcnn,thomas19kpconv,wu19pointconv,zhao19pointweb,liu19rscnn,yan20pointasnl,xiang21curvenet,zhao21pt,ma22pointmlp}, 
voxel-based~\cite{wu15modelnet,Maturana15voxnet,qi16volumetric,zhou18voxelnet}, and 
view-based methods~\cite{su15mvcnn,su18mvcnn-new,feng18gvcnn,wang19rcpcnn,Esteves19emv,ha19SeqViews2SeqLabels,
han193D2SeqViews,chen19veram,ma19learningmultiview,wei20viewgcn,wei22viewgcn++,zhang19imhl,feng19hgnn,gao23hgnn+}. 
Among them, view-based methods recognize a 3D object based on its rendered or projected images,  termed \emph{multiple~views}. 
Generally, methods in this line~\cite{su18mvcnn-new,wei20viewgcn,chen21mvt,xu21carnet} outperform the point- and voxel-based counterparts\cite{qi16volumetric,yan20pointasnl,xiang21curvenet,zhao21pt,ma22pointmlp}. 
On one hand, view-based methods benefit from massive image datasets and the advances in image recognition over the past decade. 
On the other hand, the multiple views of a 3D shape contain richer visual and semantic signals than the point or voxel form. 
For example, one may not be able to decide whether two 3D shapes 
belong to the same category by observing them from one view, but the answer becomes clear after watching other views of these shapes. 
The example inspires a central problem, \eg, how to exploit multi-view information effectively for a better understanding of 3D shape. 

\begin{figure}
   \centering
   \begin{subfigure}{0.49\linewidth}
         \includegraphics[width=\linewidth]{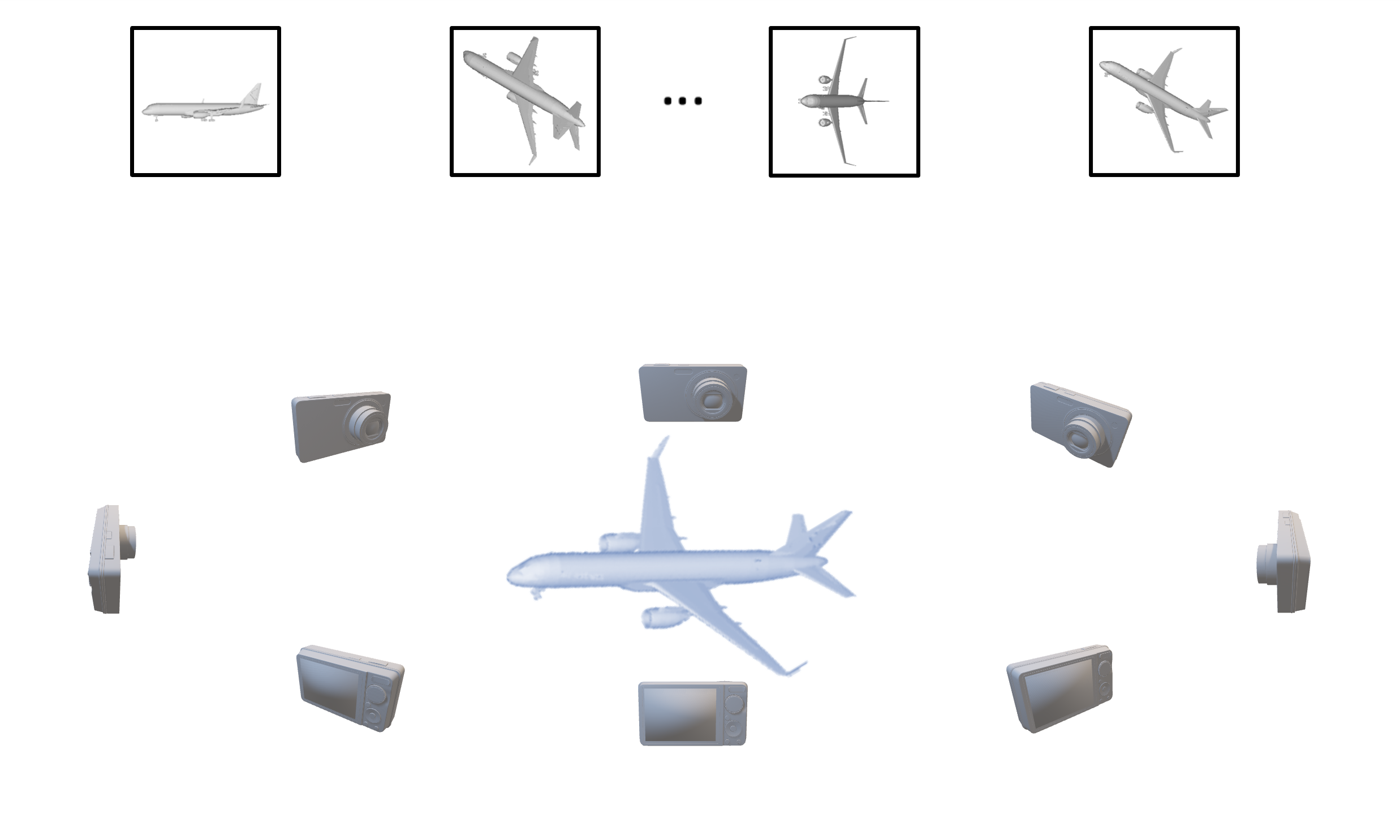}
         \caption{Independent Views}
         \label{fig:independent_view}
   \end{subfigure}
   \begin{subfigure}{0.49\linewidth}
         \includegraphics[width=\linewidth]{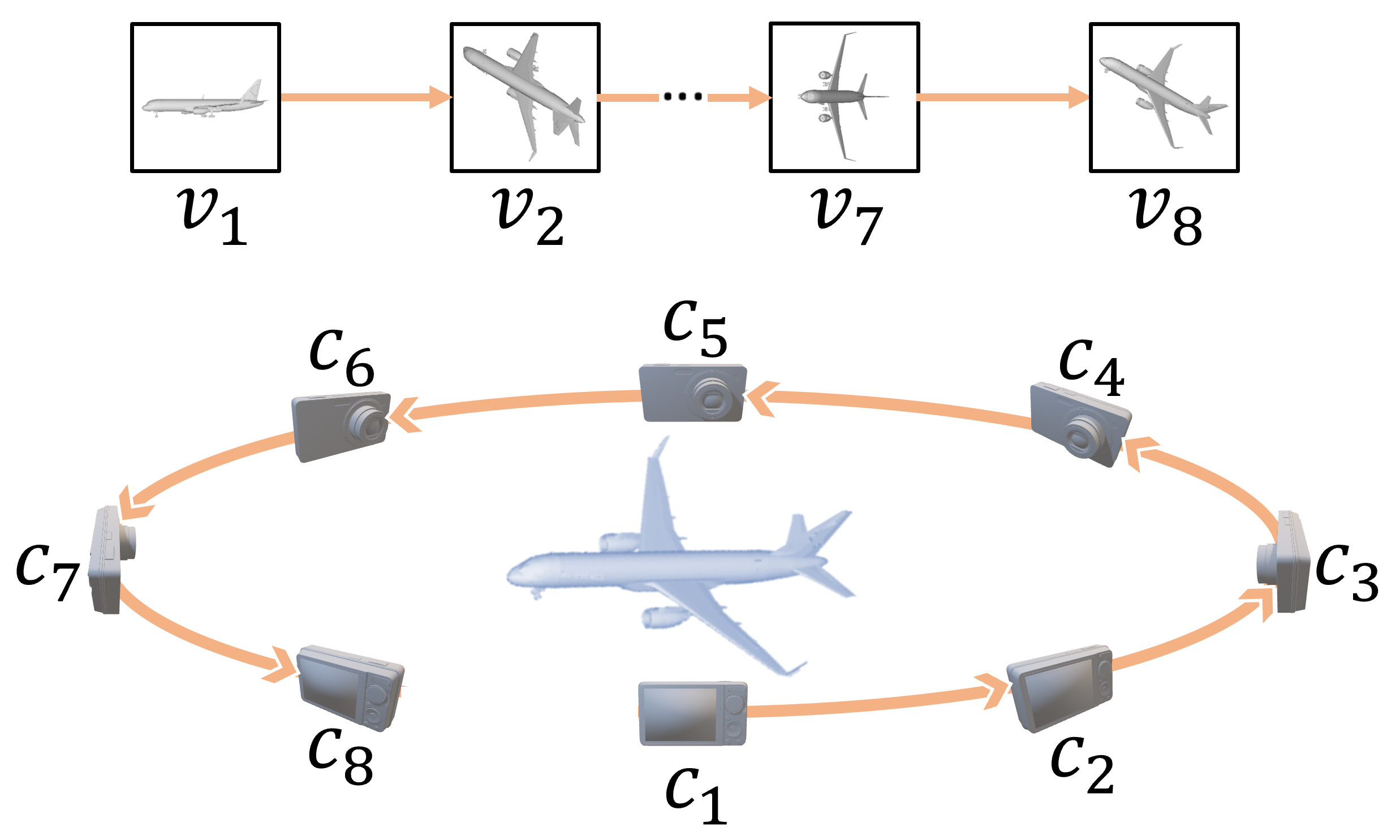}
         \caption{View Sequence}
         \label{fig:view_sequence}
   \end{subfigure}
   \begin{subfigure}{0.49\linewidth}
         \includegraphics[width=\linewidth]{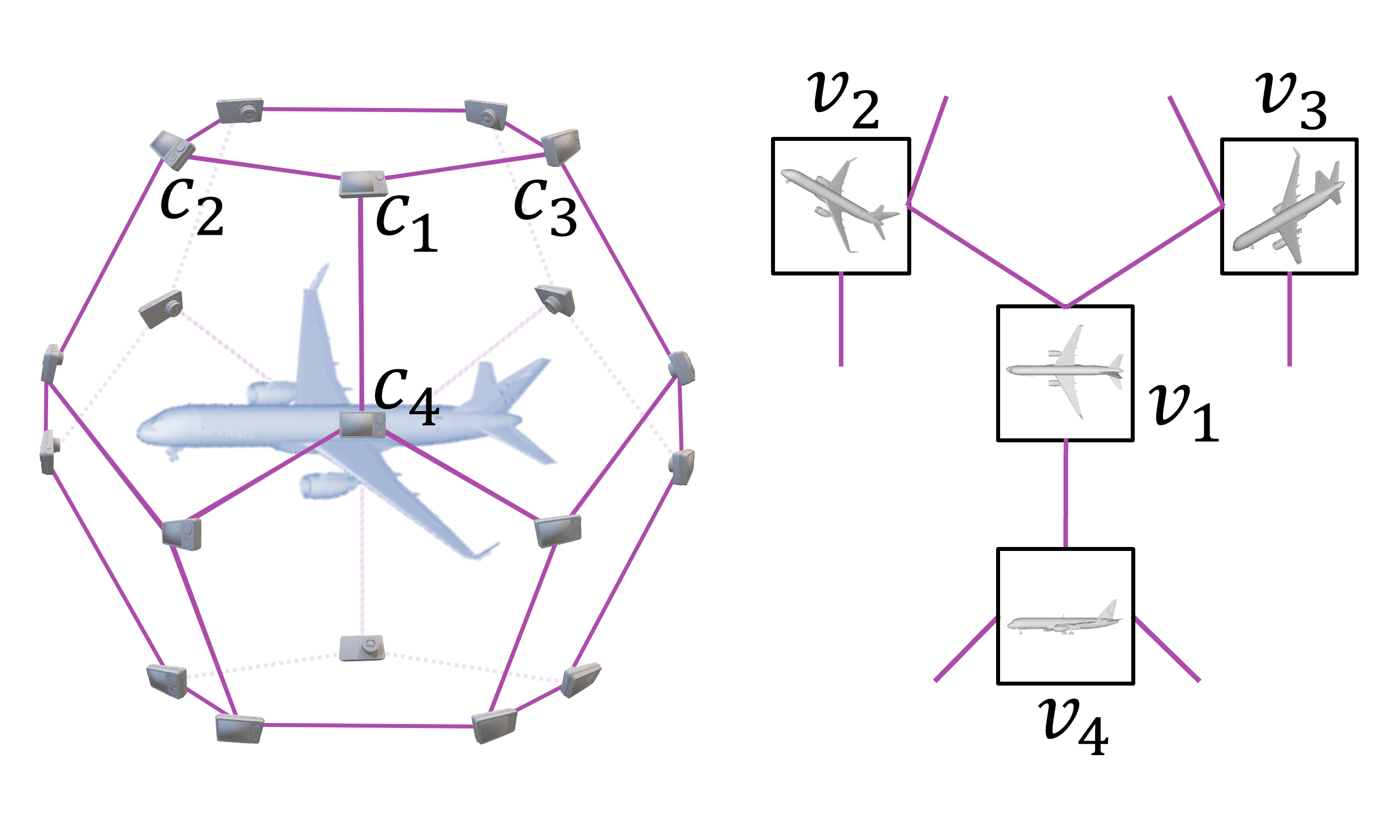}
         \caption{View Graph}
         \label{fig:view_graph}
   \end{subfigure}
   \begin{subfigure}{0.49\linewidth}
         \includegraphics[width=\linewidth]{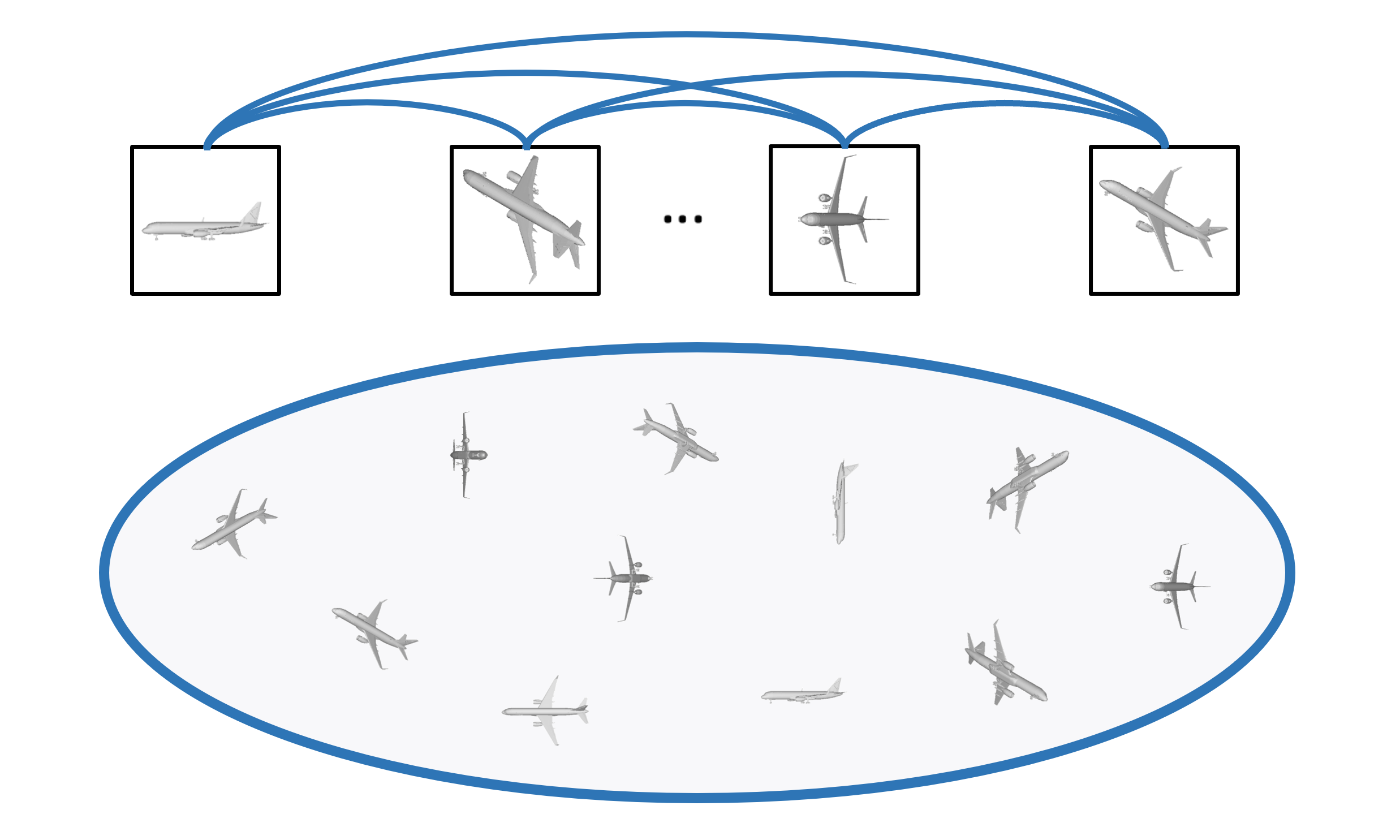}
         \caption{View Set}
         \label{fig:view_set}
   \end{subfigure}

   \caption{A division for multi-view 3D shape analysis methods based on how they organize views and 
   aggregate multi-view information. View Set is the proposed perspective that the views of a 3D shape are organized in a set.}
   \label{fig:view_structures}
\end{figure}

This paper systematically investigates existing methods on how they aggregate the multi-view information and 
the findings are summarized in Figure~\ref{fig:view_structures}. 
In the early stage, MVCNN~\cite{su15mvcnn} and its follow-up work~\cite{su18mvcnn-new,feng18gvcnn,yu18mhbn,wang19rcpcnn,yu21mhbn++} independently process multiple views of a 3D shape by a shared CNN. 
The extracted features are fused with pooling operation or some variants to form a compact 3D shape descriptor. 
We group these methods into \emph{Independent Views} in Figure~\ref{fig:independent_view}. 
Although the simple design made them stand out at the time, 
they did not take a holistic perspective to the multiple views of a 3D shape and the information flow among views was insufficient.
In the second category, a growing number of methods model multiple views as a sequence~\cite{ha19SeqViews2SeqLabels, han193D2SeqViews, chen19veram, ma19learningmultiview, yang19relationnet}, 
which are grouped into \emph{View Sequence} in Figure~\ref{fig:view_sequence}. 
They deploy RNNs, like GRU~\cite{chung14empirical} and LSTM~\cite{Hochreiter97lstm}, to learn the view relations. 
However, a strong assumption behind \emph{View Sequence} is that the views are collected from a circle around the 3D shape. 
In many cases, the assumption may be invalid since the views can be rendered from random viewpoints, so they are unordered. 
To alleviate this limitation, later methods describe views with a more general structure, 
\eg, graph~\cite{wei20viewgcn, wei22viewgcn++} or hyper-graph~\cite{zhang19imhl, feng19hgnn, gao23hgnn+}, 
and develop graph convolution networks (GCNs) to propagate and integrate view features, called \emph{View Graph} in Figure~\ref{fig:view_graph}. 
Methods in this category show both flexibility and promising performance gains, whereas they require constructing 
a view graph according to the positions of camera viewpoints. But sometimes the viewpoints may be unknown and graph construction 
introduces additional computation overheads. In addition, message propagation between remote nodes on the view graphs
may not be straightforward. 
Some other methods explore rotations~\cite{Kanezaki18rotationnet,Esteves19emv}, multi-layered height-maps representations~\cite{sarkar18mlh}, view correspondences~\cite{xu21carnet},
viewpoints selection~\cite{Hamdi21mvtn} when analyzing 3D shapes.
They can hardly be divided into the above categories, but multi-view correlations in their 
pipelines still need to be enhanced. 

By revisiting existing works, two ingredients are found critical for improving multi-view 3D shape analysis. 
The first is how to organize the views so that they can communicate with each other flexibly and freely. The second is how to 
integrate multi-view information effectively. It is worth noting that the second ingredient is usually coupled with the first, 
just like GCNs defined on the view graphs, and RNNs defined on the view sequences. 
In this paper, we present a novel perspective that multiple views of a 3D shape are organized into a \emph{View Set} in Figure~\ref{fig:view_set}, 
where elements are permutation invariant, which is consistent with the fact that 3D shape understanding is 
actually not dependent on the order of input views. For example, in Figure~\ref{fig:view_sequence}, whether the side view 
is placed first, middle or last in the inputs, the recognition result should always be \verb|airplane|. 
Unlike existing methods analyzed above, this perspective also makes no assumptions about the correlations of views, which 
is more flexible and practical in real-world applications. 
Instead, to aggregate multi-view information, a view set attention model, ViewFormer, is devised to learn the pairwise and higher-order 
relations among the views adaptively.
The attention architecture is a natural choice because it aligns with the view set characteristics. First,  
the attention mechanism is essentially a set operator and inherently good at capturing correlations between the elements in a set. 
Second, this mechanism is flexible enough that it makes minimal assumptions about the inputs, which matches our expectation
that there are no predefined relations or additional requirements for views. 

The proposed model has four components: Initializer, Encoder, Transition, and Decoder. 
Initializer initializes the representations of views. 
Encoder is adapted from standard Transformer encoder~\cite{vaswani17transformer} with specific modifications. 
i) The position encodings of input views are removed since views are permutation invariant. 
ii) The class token is removed because it is irrelevant to capturing the correlations of views in the set. 
iii) The number of attention blocks is greatly reduced as the size of a view set is relatively small ($\leq$ 20 in most cases) 
so it is unnecessary to employ deeper blocks. 
Transition summarizes the learned correlations into a compact View Set Descriptor (VSD) to express the ViewFormer's understanding of the 3D shape. 
Decoder is designed towards downstream tasks, such as recognition and retrieval. 
The simple designs around the view set show not only great flexibility but also 
powerful capability for 3D shape understanding. New records are obtained by ViewFormer in downstream tasks of 3D shape recognition and retrieval. 
In summary, the contributions of this paper include: 
\begin{itemize}
   \item A systematical investigation of existing methods in aggregating multi-views for 3D shape understanding. %
   A novel perspective is proposed that multiple views are incorporated in a \emph{View Set}. And a simple yet effective view set attention model, 
   ViewFormer, is designed to adaptively capture pairwise and higher-order correlations
   among the views for better understanding. %
   \item Extensive evaluations demonstrate the superb performances of the proposed approach.
   The recognition accuracy on ModelNet40 can reach as high as 98.8\%, %
   surpassing all existing methods. On the challenging RGBD dataset, 
   ViewFormer achieves 98.4\% classification accuracy, which is a 4.1\% absolute improvement over previous state-of-the-art. 
   ViewFormer-based 3D shape retrieval sets new records in several evaluation dimensions on SHREC'17 benchmark.
   \item Ablation studies shed light on the various sources of performance gains for 3D shape understanding and the
   visualizations provide some insightful conclusions. %
\end{itemize} 

\section{Related Work}
In this section, we review the multi-view 3D shape analysis methods and explore the deployment of set and attention in these methods. %

\noindent\textbf{Multi-view 3D Shape Analysis.} 
Existing methods aggregate multi-view information for 3D shape understanding in different ways. 
(1) \emph{Independent Views}. Early work like MVCNN series~\cite{su15mvcnn} and its follow-up~\cite{su18mvcnn-new,feng18gvcnn,yu18mhbn,wang19rcpcnn,yu21mhbn++}
extract view features independently using a shared CNN, then fuse the extracted features using the pooling operation or some variants. 
The simple strategy may discard a lot of useful information and the views are not well treated as a whole 
thus information flow among views needs to be increased. 
(2) \emph{View Sequence}. Researchers perceive the problems and propose various descriptions to incorporate multiple views 
of a 3D shape into a specific data structure. For example,
RNN-based methods~\cite{ha19SeqViews2SeqLabels, han193D2SeqViews, chen19veram, yang19relationnet, ma19learningmultiview, chen21mvt} 
are proposed to operate on the view sequence. 
(3) \emph{View Graphs}. The graph-based models~\cite{feng19hgnn, zhang19imhl, wei20viewgcn, wei22viewgcn++, gao23hgnn+} assume 
the relations among views as graphs and develop GCNs to capture multi-view interaction. 
However, message propagation on view graphs may not be straightforward and graph construction leads to additional overheads. 
(4) This paper presents a flexible and practical perspective, \emph{View Set}, which neither makes
assumptions about views nor introduces additional overheads. 
Based on that, a view set attention model is devised to adaptively integrate the correlations for all view pairs. 
Some other methods also explore rotations~\cite{Kanezaki18rotationnet,Esteves19emv}, multi-layered height-maps representations~\cite{sarkar18mlh}, view correspondences~\cite{xu21carnet},
viewpoints selection~\cite{Hamdi21mvtn} when analyzing 3D shapes. Their multi-view interaction still needs to be strengthened. 

\noindent\textbf{Set in Multi-view 3D Shape Analysis.} 
Previous works also mention ``set" in multi-view 3D shape analysis. 
But they basically refer to different concepts from the proposed one. 
For instance, RCPCNN~\cite{wang19rcpcnn} introduces a dominant set clustering and pooling
module to improve MVCNN~\cite{su15mvcnn}.
Johns \etal decompose a sequence of views into a set of view pairs. They classify each pair independently and weigh the 
contribution of each pair~\cite{johns16pairwise}. MHBN~\cite{yu18mhbn} considers patches-to-patches (set-to-set) similarity of different views
and aggregates local features using bilinear pooling. Yu \etal extend MHBN by introducing VLAD layer~\cite{yu21mhbn++}. 
The basic idea is to calculate the similarity between two sets of local patches, while our view set idea provides a foundation for 
adaptively learning inter-view attentions. 

\noindent\textbf{Attention in Multi-view 3D Shape Analysis.} 
The attention mechanisms have been embedded in existing multi-view 3D shape recognition methods, but they vary in 
motivation, practice and effectiveness. VERAM~\cite{chen19veram} uses a recurrent attention model to select
a sequence of views to classify 3D shapes adaptively. SeqViews2SeqLabels~\cite{ha19SeqViews2SeqLabels} introduces the attention 
mechanism to increase the discriminative ability for the RNN-based model and reduces the effect of selecting the first view position.  
3D2SeqViews~\cite{han193D2SeqViews} proposes hierarchical attention to incorporate view-level and class-level importance
for 3D shape analysis. Nevertheless, there are two points worth noting for the attention of the above methods. 
First, the attention operation in these
methods differs from multi-head self-attention in standard Transformer~\cite{vaswani17transformer}. Second, 
the dedicated designed attention does not seem to produce satisfactory results since the highest recognition accuracy 
they achieve on ModelNet40 is 93.7\%, whereas our solution reaches 99.0\% on the same dataset. 
Recent work MVT~\cite{chen21mvt} also explores the attention architecture for view-based 3D recognition. 
It is inspired by the success of ViT~\cite{dosovitskiy21vit} in image recognition and 
wants to strengthen view-level communications with patch-level correlations. 
MVT deploys a ViT to extract patch-level features for all images and adopts another ViT to learn the correlations for all views. 
However, ViewFormer shows it is unnecessary to take the patch-level interactions into account to achieve the best results, 
thus the computation budgets are considerably reduced compared to MVT. 

\section{ViewFormer}
\label{sec:method}

In this section, we firstly formulate the problem of multi-view 3D shape recognition and retrieval 
based on the view set, then elaborate on the devised model and how it handles a set of views. 

\subsection{Problem Formulation}

\noindent\textbf{View Set.} The views of a 3D shape refer to the rendered or projected RGB images from it. For example, 
a 3D shape $\mathcal{S}$ corresponds to views $v_1, v_2, \dots, v_M \in \mathbb{R}^{H\times W\times C}$, where $M$ is the number of views 
and $H\times W\times C$ indicates the image size. 
In our perspective, the views of $\mathcal{S}$ simply form a set $\mathcal{V} = \{v_1, v_2, \dots, v_M\}$, 
where elements are permutation invariant. Thus $\mathcal{V}$ can be instantiated as a random permutation of the views. 
This perspective matches the basic fact that views can be generated from random viewpoints in the real world. 
It neither assumes relations for views nor introduces additional overheads, distinguished from previous methods analyzed above. 

\noindent\textbf{3D Shape Recognition \& Retrieval.}
In many cases~\cite{savva17shrec17}, 3D shape retrieval can be regarded as a classification problem. It aims to find the most relevant shapes to the query. Meanwhile, the relevance is defined according to the ground truth class and subclass of the query, which means if a retrieved shape has the same class and subclass as the query, they match perfectly. Therefore, the tasks of 3D shape recognition and retrieval can be unified by 
predicting a category distribution $\hat{\textbf{y}} \in \mathbb{R}^{K}$ of the target shape $\mathcal{S}$, where $K$ is the number of 3D shape categories. 
In this paper, we design a simple yet effective view set attention model $\mathcal{F}$ to 
predict the distribution. 
The input of $\mathcal{F}$ is a view set $\mathcal{V} \in \mathbb{R}^{M\times H\times W\times C}$, corresponding to the shape $\mathcal{S}$. 
The procedure is formulated by Eq.~\ref{eq:3d_shape_recog} and the details are dissected in the next section. 
\begin{equation}
   \hat{\textbf{y}} = \mathcal{F}(\mathcal{V})
   \label{eq:3d_shape_recog}
\end{equation}

\subsection{View Set Attention Model}
The proposed view set attention model, ViewFormer, is to 
adaptively grasp pairwise and higher-order correlations among views in the set. 
And it summarizes the learned correlations into an expressive descriptor for 3D shape analysis. 
ViewFormer is more straightforward in modeling the correlations of views 
than graph-based methods because it explicitly computes the attention scores for all view pairs. 
The overall architecture of 
ViewFormer includes four modules: Initializer, Encoder, Transition, and Decoder. 

\noindent\textbf{Initializer.} %
This module initializes the feature representations of views in $\mathcal{V}$ to feed Encoder. 
We denote the module as Init and it converts $v_i \in \mathbb{R}^{H\times W\times C}$ to the 
feature representation $z_i \in \mathbb{R}^D$, where $D$ is the feature dimension. 
After this module, the view set $\mathcal{V} = \{v_1, \dots, v_i, \dots, v_M\}$ 
is transformed to the initialized feature set $\textbf{z}^0 = \{z_1, \dots, z_i, \dots, z_M\}$, shown in Eq.~\ref{eq:init}. 

\begin{equation}
	\textbf{z}^0 = \textrm{Init}(\mathcal{V})
	\label{eq:init}
\end{equation}

Init has various choices, such as linear projection, MLP, CNN or ViT. The complexity and efficiency are tradeoffs. 
A simple linear projection from a $224\times 224\times 3$ view to a 512-dimensional vector will result in 
$\sim$77M parameters in Init, and the MLP will produce more. 
Some work~\cite{yu18mhbn,yu21mhbn++,chen21mvt} also consider fine-grained patch-level features 
within each view and then combine them with the view-level ones. %
But this mean is computation expensive. 
In ViewFormer, we adopt lightweight CNNs (\eg, AlexNet~\cite{krizhevsky12alexnet}, ResNet18~\cite{he2016resnet}) 
as Init because they are efficient and good at image feature extraction.

\noindent\textbf{Encoder.} 
This module that consists of consecutive attention blocks is 
adapted from standard Transformer~\cite{vaswani17transformer} encoder with the following modifications. 
First, the position encodings are removed since the views should be unaware of their order in the view set. 
Second, the class token is removed because it is irrelevant to the target of modeling the correlations of views in the set. 
Third, the number of attention blocks is greatly reduced as the size of a view set is relatively small 
($\leq$ 20 in most cases), so employing very complex encoder is inappropriate. 

Encoder receives the initialized view feature set $\textbf{z}^0 \in \mathbb{R}^{M\times D}$ and processes them with $L$ 
attention blocks. Each attention block stacks the multi-head self-attention\cite{vaswani17transformer} (MSA) and MLP layers with residual connections. 
LayerNorm (LN) is deployed before MSA and MLP, whereas Dropout is applied after them. 
The feature interaction is explicitly calculated for all view pairs in each attention block and 
by going deeper, the higher-order correlations are learned. 
The procedure in the $\ell$th block is summarized by Eq.~\ref{eq:msa} and~\ref{eq:mlp}. 
\begin{equation}
	\hat{\textbf{z}}^{\ell} = \textrm{Dropout}(\textrm{MSA}(\textrm{LN}(\textbf{z}^{\ell-1}))) + \textbf{z}^{\ell-1} \quad \ell=1\dots L
	\label{eq:msa}
\end{equation}
\begin{equation}
	\textbf{z}^{\ell} = \textrm{Dropout}(\textrm{MLP}(\textrm{LN}(\hat{\textbf{z}}^{\ell}))) + \hat{\textbf{z}}^{\ell} \quad \ell=1\dots L
	\label{eq:mlp}
\end{equation}

\noindent\textbf{Transition.} 
The last attention block of Encoder outputs the collective correlations of multiple views $\textbf{z}^L \in \mathbb{R}^{M\times D}$ and 
we convert the learned correlations into a view set descriptor by the Transition module (Transit). 
The pooling operations are typical options in existing methods~\cite{su15mvcnn,wang19rcpcnn,feng18gvcnn,yu18mhbn,yu21mhbn++}. 
In this paper, we concatenate (Concat) the results of max and mean pooling along the first dimension of $\textbf{z}^L$
to stabilize the optimization and the operation does not introduce learnable parameters. 
The output is denoted as $\textbf{t}^L \in \mathbb{R}^{2D}$ in Eq.~\ref{eq:transition}. 
\begin{equation}
		\textbf{t}^L = \textrm{Transit}(\textbf{z}^L) 
		= \textrm{Concat}(\textrm{Max}(\textbf{z}^L), \textrm{Mean}(\textbf{z}^L))
	\label{eq:transition}
\end{equation}

\noindent\textbf{Decoder.} 
This module decodes the view set descriptor $\textbf{t}^L$ to a 3D shape category distribution $\hat{\textbf{y}} \in \mathbb{R}^{K}$. 
In ViewFormer, we show the decoder can be designed extremely lightweight, as light as a single Linear. 
We also make a look into the performance of heavier heads, such as 2- or 3-layer MLP preceded by BatchNorm (BN) and ReLU in each layer. 
We find both of them work well, reflecting the summarized view set descriptor $\textbf{t}^L$ is highly expressive. 
\begin{equation}
	\hat{\textbf{y}} = \textrm{Decoder}(\textbf{t}^L)
	\label{eq:decoder}
\end{equation}

By combining the simple design of each component, the proposed method exhibits powerful capabilities across
different datasets and tasks, supported by systematic experiments and extensive ablation studies in the next section.

\section{Experiments}
\label{sec:experiments}
In this section, firstly, we explain the experimental settings of ViewFormer. 
Then the proposed method is evaluated on 3D shape recognition and retrieval tasks. 
Thirdly, we conduct controlled experiments to justify the design choices of ViewFormer. 
Finally, visualizations are presented for a better understanding of the method. 

\begin{table}
   \begin{center}
      \begin{tabular}{l l c c}
      \toprule
      \multirow{2}{*}{Method} & \multirow{2}{*}{Input} & Class Acc. & Inst. Acc.\\
      & & (\%) & (\%)\\
      \midrule
      3DShapeNets~\cite{wu15modelnet} & \multirow{4}{*}{Voxels} & 77.3 & -- \\
      VoxNet~\cite{Maturana15voxnet} &  & 83.0 & -- \\
      VRN Ensemble~\cite{brock16vrn} &  & -- & 95.5 \\
      MVCNN-MR~\cite{qi16volumetric} &  & 91.4 & 93.8 \\\hdashline
      PointNet++~\cite{qi17pointnet2} & \multirow{6}{*}{Points} & -- & 91.9 \\
      DGCNN~\cite{wang19dgcnn} &  & 90.2 & 92.9 \\
      RSCNN~\cite{liu19rscnn} &  & -- & 93.6 \\
      KPConv~\cite{thomas19kpconv} &  & -- & 92.9 \\
      CurveNet~\cite{xiang21curvenet} &  & -- & 93.8 \\
      PointMLP~\cite{ma22pointmlp} &  & 91.3 & 94.1 \\\hdashline
      MVCNN~\cite{su15mvcnn} & \multirow{21}{*}{Views} & 90.1 & 90.1 \\
      MVCNN-new~\cite{su18mvcnn-new} &  & 92.4 & 95.0 \\
      MHBN~\cite{yu18mhbn} &  & 93.1 & 94.7 \\
      GVCNN~\cite{feng18gvcnn} &  & 90.7 & 93.1 \\
      RCPCNN~\cite{wang19rcpcnn} &  & -- & 93.8 \\
      RN~\cite{yang19relationnet} & & 92.3 & 94.3 \\
      3D2SeqViews~\cite{han193D2SeqViews} &  & 91.5 & 93.4 \\
      SV2SL~\cite{ha19SeqViews2SeqLabels} &  & 91.1 & 93.3 \\
      VERAM~\cite{chen19veram} &  & 92.1 & 93.7 \\
      Ma \etal~\cite{ma19lmvr} &  & -- & 91.5 \\
      iMHL~\cite{zhang19imhl} &  & -- & 97.2 \\
      HGNN~\cite{feng19hgnn} &  & -- & 96.7 \\
      HGNN$^+$~\cite{gao23hgnn+} &  & -- & 96.9 \\
      View-GCN~\cite{wei20viewgcn} &  & \textcolor{blue}{96.5} & 97.6 \\
      View-GCN++~\cite{wei22viewgcn++} &  & \textcolor{blue}{96.5} & 97.6 \\
      DeepCCFV~\cite{huang19deepccfv} &  & -- & 92.5 \\
      EMV~\cite{Esteves19emv} &  & 92.6 & 94.7 \\
      RotationNet~\cite{Kanezaki18rotationnet} &  & -- & 97.4 \\
      MVT~\cite{chen21mvt} &  & -- & 97.5 \\
      CARNet~\cite{xu21carnet} &  & -- & \textcolor{blue}{97.7} \\
      MVTN~\cite{Hamdi21mvtn} &  & 92.2 & 93.5 \\
      \midrule
      \textbf{ViewFormer} & Views & \textbf{98.9} & \textbf{98.8}\\
      \bottomrule
      \end{tabular}
      \caption{Comparison of 3D shape recognition on ModelNet40. The best score is in bold black
      and the second best is in blue. The convention is kept in the following tables.}
      \label{tab:recog_modelnet40}
   \end{center}
\end{table}

\subsection{Basic Configurations}
\label{subsec:exp_config}

\noindent\textbf{Architecture.}
For Initializer, we adopt lightweight CNNs. There are several candidates (AlexNet, ResNet18, \etc) and we will compare them later. 
The view $z_i \in \mathcal{V}$ is mapped to a 512-dimensional vector through Initializer. 
For Encoder, there are $L$=4 attention blocks and within each block, the MSA layer has 8 attention heads and the widening
factor of the MLP hidden layer is 2. 
The Transition module converts the collective correlations in $\textbf{z}^L$ into a 1024-dimensional descriptor. 
Finally, the descriptor is projected to a category distribution by Decoder, which is a 2-layer MLP of 
shape \{1024, 512, $K$\}. The design choices are verified by ablated studies in Section~\ref{subsec:exp_ablation}. 

\begin{table}
   \begin{center}
      \begin{tabular}{l l c c}
      \toprule
      \multirow{2}{*}{Method} & \multirow{2}{*}{Input} & Class Acc. & Inst. Acc.\\
      & & (\%) & (\%)\\
      \midrule
      3D2SeqViews~\cite{han193D2SeqViews} & \multirow{6}{*}{Views} & 94.7 & 94.7 \\
      SV2SL~\cite{ha19SeqViews2SeqLabels} &  & 94.6 & 94.7 \\
      VERAM~\cite{chen19veram} &  & \textcolor{blue}{96.1} & 96.3 \\
      RotationNet~\cite{Kanezaki18rotationnet} &  & -- & 98.5 \\
      CARNet~\cite{xu21carnet} &  & -- & 99.0 \\
      MVT~\cite{chen21mvt} &  & -- & \textcolor{blue}{99.3} \\
      \midrule
      \textbf{ViewFormer} & Views & \textbf{100.0} & \textbf{100.0}\\
      \bottomrule
      \end{tabular}
      \caption{Comparison of 3D shape recognition on ModelNet10.}
      \label{tab:recog_modelnet10}
   \end{center}
\end{table}

\noindent\textbf{Optimization.}
The loss function is defined as CrossEntropyLoss for 3D shape recognition. Following previous methods~\cite{su18mvcnn-new,wei20viewgcn}, 
the learning is divided into two stages. In the first stage, the Initializer is individually trained on the target dataset
for 3D shape recognition. The purpose is to provide good initializations for views.  
In the second stage, the pre-trained Initializer is loaded and jointly optimized with other modules on the same dataset. 
Experiments show this strategy will significantly improve performance in a shorter period. 
More explanations about network optimization and evaluations of learning efficiency are provided 
in the supplementary material.

\subsection{3D Shape Recognition}
\label{subsec:exp_recognition}

\noindent\textbf{Datasets \& Metrics.} 
We conduct 3D shape recognition on three datasets, ModelNet10~\cite{wu15modelnet}, ModelNet40~\cite{wu15modelnet} and RGBD~\cite{lai11rgbd}. 
ModelNet10 has 4,899 CAD models in 10 categories %
and ModelNet40 includes 12,311 objects across 40 categories. %
For ModelNet10/40, we use their rendered versions as in previous work~\cite{su18mvcnn-new,wei20viewgcn}, 
where each object corresponds to 20 views. 
RGBD is a large-scale, hierarchical multi-view object dataset~\cite{lai11rgbd}, containing 300 objects organized into 51 classes. 
In RGBD, we use 12 views for each 3D object as in~\cite{Kanezaki18rotationnet,wei20viewgcn}. 
Two evaluation metrics are computed for 3D shape recognition: mean class accuracy (Class Acc.) and instance accuracy (Inst. Acc.). 
We record the best results of these metrics during optimization. 

\begin{table}
   \begin{center}
      \begin{tabular}{l c c}
      \toprule
      Method & \#Views & Inst. Acc. (\%)\\
      \midrule
      CFK~\cite{cheng15cfk} & $\geq$ 120 & 86.8\\
      MMDCNN~\cite{rahman17rgbdor} & $\geq$ 120 & 89.5\\
      MDSICNN~\cite{asif18amm} & $\geq$ 120 & 89.6\\
      MVCNN~\cite{su15mvcnn} & 12 & 86.1 \\
      RotationNet~\cite{Kanezaki18rotationnet} & 12 & 89.3 \\\hdashline
      View-GCN(ResNet18)~\cite{wei20viewgcn} & 12 & \textcolor{blue}{94.3} \\
      View-GCN(ResNet50)~\cite{wei20viewgcn} & 12 & \textcolor{blue}{93.9} \\
      \midrule
      \textbf{ViewFormer}(ResNet18) & 12 & \textbf{98.4} \\
      \textbf{ViewFormer}(ResNet50) & 12 & \textbf{95.6} \\
      \bottomrule
      \end{tabular}
      \caption{Comparison of 3D shape recognition on RGBD.}
      \label{tab:recog_rgbd}
   \end{center}
\end{table}

\begin{table*}[ht]
	\begin{center}
		\begin{tabular}{l c c c c c c c c c c c}
			\toprule
			\multirow{2}{*}{Method} & \multicolumn{5}{c}{micro} & \multicolumn{5}{c}{macro} \\
			\cline{2-6}                     \cline{8-12}
			& P@N & R@N & F1@N & mAP & NDCG & & P@N & R@N & F1@N & mAP & NDCG\\
			\midrule
			ZFDR & 53.5 & 25.6 & 28.2 & 19.9 & 33.0 & & 21.9 & 40.9 & 19.7 & 25.5 & 37.7 \\
			DeepVoxNet & 79.3 & 21.1 & 25.3 & 19.2 & 27.7 & & 59.8 & 28.3 & 25.8 & 23.2 & 33.7 \\
			DLAN & \textbf{81.8} & 68.9 & 71.2 & 66.3 & 76.2 & & 61.8 & 53.3 & 50.5 & 47.7 & 56.3 \\\hdashline
			GIFT~\cite{bai16gift} & 70.6 & 69.5 & 68.9 & 64.0 & 76.5 & & 44.4 & 53.1 & 45.4 & 44.7 & 54.8 \\
			Improved GIFT~\cite{bai17gift} & 78.6 & 77.3 & 76.7 & 72.2 & 82.7 & & 59.2 & 65.4 & 58.1 & 57.5 & 65.7 \\
			ReVGG & 76.5 & 80.3 & 77.2 & 74.9 & 82.8 & & 51.8 & 60.1 & 51.9 & 49.6 & 55.9 \\
			MVFusionNet & 74.3 & 67.7 & 69.2 & 62.2 & 73.2 & & 52.3 & 49.4 & 48.4 & 41.8 & 50.2 \\
			CM-VGG5-6DB & 41.8 & 71.7 & 47.9 & 54.0 & 65.4 & & 12.2 & \textbf{66.7} & 16.6 & 33.9 & 40.4 \\
			MVCNN~\cite{su15mvcnn} & 77.0 & 77.0 & 76.4 & 73.5 & 81.5 & & 57.1 & 62.5 & 57.5 & 56.6 & 64.0 \\
			RotationNet~\cite{Kanezaki18rotationnet} & 81.0 & 80.1 & 79.8 & 77.2 & \textbf{86.5} & & 60.2 & 63.9 & 59.0 & 58.3 & 65.6 \\
			View-GCN~\cite{wei20viewgcn} & \textbf{81.8} & \textcolor{blue}{80.9} & \textcolor{blue}{80.6} & \textbf{78.4} & \textcolor{blue}{85.2} & & \textcolor{blue}{62.9} & 65.2 & \textcolor{blue}{61.1} & \textcolor{blue}{60.2} & \textcolor{blue}{66.5} \\
			View-GCN++~\cite{wei22viewgcn++} & 81.2 & 79.9 & 80.0 & \textcolor{blue}{77.5} & 83.9 & & 61.2 & \textcolor{blue}{65.8} & \textcolor{blue}{61.1} & 59.0 & 63.8 \\
			\midrule
			$\textbf{ViewFormer}$ & \textcolor{blue}{81.6} & \textbf{82.0} & \textbf{81.3} & \textbf{78.4} & 81.7 & & \textbf{64.5} & 65.4 & \textbf{62.9} & \textbf{60.6} & \textbf{67.5} \\ 
			\bottomrule
		\end{tabular}
		\caption{Comparison of 3D shape retrieval on ShapeNet Core55.}
		\label{tab:ret_shrec17}
	\end{center}
\end{table*}

\noindent\textbf{Results.} 
Table~\ref{tab:recog_modelnet40} compares representative methods on ModelNet40 and these methods have different input formats: 
voxels, points and views. 
ViewFormer achieves 98.9\% mean class accuracy and 98.8\% overall accuracy, surpassing the voxel- and point-based counterparts.  
Also, it sets new records in view-based methods. For example, compared to early works~\cite{su15mvcnn,su18mvcnn-new,yu18mhbn,feng18gvcnn,wang19rcpcnn}
that aggregate multi-view information independently by pooling or some variants,  
ViewFormer exceeds their instance accuracies by 3.8\% at least. 
ViewFormer also significantly improves the results of methods built on view sequence, such as RelationNet~\cite{yang19relationnet}, 
3D2SeqViews~\cite{han193D2SeqViews}, SeqViews2SeqLabels~\cite{ha19SeqViews2SeqLabels}, VERAM~\cite{chen19veram}. 
Methods defined on view graph and hyper-graph achieve decent performances~\cite{zhang19imhl,feng19hgnn,gao23hgnn+,wei20viewgcn,wei22viewgcn++} 
because of enhanced information flow among views. ViewFormer still outreaches the strongest baseline of this category,  
increasing 2.4\% Class Acc. and 1.2\% Inst Acc. over View-GCN~\cite{wei20viewgcn}. 

Table~\ref{tab:recog_modelnet10} presents the recognition results on ModelNet10. 
Although the dataset is relatively easy and previous methods can work very well (as high as 99.3\% Inst. Acc.), 
it is a bit surprising that ViewFormer successfully recognizes all shapes in the test set and obtains 100\% accuracy. 
Previous best method MVT~\cite{chen21mvt} combines patch- and view-level feature communications by applying 
ViT~\cite{dosovitskiy21vit} twice. ViewFormer achieves better results without taking patch-level interaction into account. 

Table~\ref{tab:recog_rgbd} records the comparison with related work on the challenging RGBD~\cite{lai11rgbd} dataset. 
The dataset designs 10-fold cross-validation for multi-view 3D object recognition. 
We follow this setting and report the average instance accuracy of 10 folds. 
ViewFormer shows consistent improvements over View-GCN under the same Initializer. Especially, 
it gets 98.4\% accuracy that is a 4.1\% absolute improvement over the runner-up, 
suggesting ViewFormer can produce more expressive shape descriptors when dealing with challenging cases. 

\subsection{3D Shape Retrieval}
\label{subsec:exp_retrieval}

\noindent\textbf{Datasets \& Metrics.}
3D shape retrieval aims to find a rank list of shapes most relevant to the query shape in a given dataset. 
We conduct this task on ShapeNet Core55~\cite{shapenet2015,savva17shrec17}. The dataset is split into train/val/test 
set and there are 35764, 5133 and 10265 meshes, respectively. 
20 views are rendered for each mesh as in~\cite{Kanezaki18rotationnet,wei20viewgcn}. 
According to the SHREC'17 benchmark~\cite{savva17shrec17}, the rank list is evaluated based on the ground truth 
category and subcategory. If a retrieved shape in a rank list has the same category as the query, it is positive. Otherwise, it is negative. 
The evaluation metrics include micro and macro version of P@N, R@N, F1@N, mAP and NDCG. 
Here N is the length of returned rank list and its maximum value is 1,000 according to the requirement. 
Please refer to \cite{savva17shrec17} for more details about the metrics. 

\noindent\textbf{Retrieval.} 
We generate the rank list for each query shape in two steps. 
First, ViewFormer is trained to recognize the shape categories in ShapeNet Core55~\cite{shapenet2015}. 
We retrieve shapes that have the same predicted class as the query $\mathcal{Q}$ and rank the retrieved shapes according to 
class probabilities in descending order, resulting in L$_1$.
Second, we train another ViewFormer to recognize the shape subcategories of ShapeNet Core55~\cite{shapenet2015}, 
then re-rank L$_1$ to ensure shapes that 
have same predicted subcategory as the query $\mathcal{Q}$ rank before shapes that are not in same subcategory with $\mathcal{Q}$
and keep the remaining unchanged, 
resulting in L$_2$, which is regarded as the final rank list for the query $\mathcal{Q}$. 

\noindent\textbf{Results.} 
ViewFormer is compared with the methods that report results on SHREC'17 benchmark~\cite{savva17shrec17}, 
shown in Table~\ref{tab:ret_shrec17}. 
The methods in the first three rows use voxel representations of 3D shapes as inputs, while the remaining methods exploit multiple views. 
The overall performances of view-based methods are better than voxel-based ones.  
Previously, View-GCN achieved state-of-the-art results by enhancing view interaction and aggregating multi-view information on
on view-graphs. 
But experiments show ViewFormer goes beyond View-GCN in terms of micro-version R@N, F1@N and mAP as well as 
macro-version P@N, F1@N, mAP and NDCG. 
For example, we achieve at least 1.0\% absolute improvements for both micro-version R@N and macro-version NDCG over View-GCN. 

\subsection{Ablation Studies}
\label{subsec:exp_ablation}
We conduct a series of controlled experiments to verify the choices in ViewFormer design. 
The used dataset is ModelNet40. 

\noindent\textbf{Initializer.}
We explore different means to initialize view representations, 
including shallow convolution operations and lightweight CNNs. 
The idea of shallow convolution operation is inspired by the image patch projection (1x1 Conv) in ViT~\cite{dosovitskiy21vit} and 
the specific configurations are explained in the supplementary material. 
Table~\ref{tab:ablate_view_init} compares their recognition accuracies. 
We observe that initializations by 1- and 2-layer convolution operations do not yield satisfactory
results. Instead, 
lightweight CNNs work well, especially when receiving the initialized features by AlexNet and jointly optimizing with other modules, 
ViewFormer reaches 98.9\% class accuracy and 98.8\% overall accuracy, both are new records on ModelNet40. 
By default, AlexNet serves as the Initializer module.
\begin{table}[ht]
	\begin{center}
		\begin{tabular}{l r c c}
			\toprule
			\multirow{2}{*}{Initializer} & \#Params &Class Acc.& Inst. Acc. \\
			& (M) & (\%) & (\%) \\
			\midrule
			1-layer Conv & 102.8 & 90.1 & 92.5 \\  %
			2-layer Conv & 12.9 & 88.9 & 93.7 \\\hdashline  %
			alexnet & 42.3 & \textbf{98.9} & \textbf{98.8} \\ %
			resnet18 & 11.2 & 96.7 & \textcolor{blue}{97.6}\\
			resnet34 & 21.3 & \textcolor{blue}{96.9} & 97.1 \\
			\bottomrule
		\end{tabular}
		\caption{Ablation study: choices for Initializer.}
		\label{tab:ablate_view_init}
	\end{center}
\end{table}

\noindent\textbf{Position Encoding.}
According to the view set perspective, 
ViewFormer should be unaware of the order of elements in the view set so we remove the position 
encoding from the devised encoder. We examine this design in Table~\ref{tab:ablate_pos_enc}.
The results show 
if learnable position embeddings are forcibly injected into the initialized view features to make 
the model position-aware, the performance will be hindered, dropping by 0.5\% for class accuracy
and 0.3\% for overall accuracy. 
\begin{table}[ht]
	\begin{center}
		\begin{tabular}{l c c}
			\toprule
			Variants & Class Acc. (\%) & Inst. Acc. (\%) \\
			\midrule
			w/ pos. enc. & \textcolor{blue}{98.4} & \textcolor{blue}{98.5}\\
			w/o pos. enc. & \textbf{98.9} & \textbf{98.8}\\
			\midrule
			w/ cls. token & \textcolor{blue}{98.8} & \textcolor{blue}{98.5} \\
			w/o cls. token & \textbf{98.9} & \textbf{98.8}\\
			\bottomrule
		\end{tabular}
		\caption{Ablation study: position encoding and class token.}
		\label{tab:ablate_pos_enc}
	\end{center}
\end{table}

\noindent\textbf{Class Token.} Unlike standard Transformer~\cite{vaswani17transformer}, 
the proposed method does not insert the class token into the inputs since 
it is irrelevant to the target of capturing the correlations among views in the set. 
This claim is supported by the results in Table~\ref{tab:ablate_pos_enc}, which shows that 
inserting the class token results in decreasing recognition accuracies. 

\noindent\textbf{Number of Attention Blocks.} 
In ViewFormer, the number of attention blocks in Encoder is considerably compressed because 
the size of a view set is relatively small and it is unnecessary to deploy a deeper encoder to model 
the interactions between the views in the set. The results in Table~\ref{tab:ablate_num_attn_blocks}
demonstrate the encoder can be highly lightweight, as light as two attention blocks, but with outstanding 
performance compared to existing methods. 
The results also indicate 
increasing the attention blocks does not receive gains but additional parameters and overheads. 
\begin{table}
	\begin{center}
		\begin{tabular}{l r r}
			\toprule
			Module & \#Params (M) & Inst. Acc. (\%) \\
			\midrule
			AlexNet & 42.3 & 85.1 \\
			\quad + 2 Attn. Blocks & \textbf{4.8} & \textbf{98.8} \\
			\quad + 4 Attn. Blocks & \textcolor{blue}{9.0} & \textbf{98.8} \\
			\quad + 6 Attn. Blocks & 13.2 & \textcolor{blue}{98.3} \\
			\bottomrule
		\end{tabular}
		\caption{Ablation study: number of attention blocks.}  %
		\label{tab:ablate_num_attn_blocks}
	\end{center}
\end{table}

\noindent\textbf{Transition.} 
We investigate three kinds of operations for the Transition module. The results are reported 
in Table~\ref{tab:ablate_transition_module}. We find the simple pooling operations (Max and Mean) can work well (98.0+\% Acc.) and
both outreach the performances of previous state of the art. By concatenating the outputs of max and mean pooling, 
the optimization is more stable and the overall accuracy is lifted to 98.8\%. 
It is worth noting that the same pooling operations are adopted by MVCNN~\cite{su15mvcnn} and its variants~\cite{su18mvcnn-new,feng18gvcnn,yu18mhbn,wang19rcpcnn,yu21mhbn++}, 
but their accuracies are up to 95.0\%, implying that the view set descriptors learned by our encoder are more informative. 
\begin{table}[ht]
	\begin{center}
		\begin{tabular}{l c c}
			\toprule
			Transition & Class Acc. (\%) & Inst. Acc. (\%) \\
			\midrule
			Max pooling & \textbf{99.1} & \textcolor{blue}{98.5} \\
			Mean pooling & 98.5 & \textcolor{blue}{98.5} \\
			Concat(Max\&Mean) & \textcolor{blue}{98.9} & \textbf{98.8} \\
			\bottomrule
		\end{tabular}
		\caption{Ablation study: choices for Transition.}
		\label{tab:ablate_transition_module}
	\end{center}
\end{table}

\noindent\textbf{Decoder.} The decoder projects the view set descriptor to a shape category distribution. 
The choices for the decoder are compared in Table~\ref{tab:ablate_decoder}. 
ViewFormer with a decoder of a single Linear can recognize 3D shapes at 98.1\% instance accuracy, 
which outperforms all existing methods and again, reflects the summarized view set descriptor is highly discriminative. 
The advantage is enlarged when the decoder is deepened to a 2-layer MLP. However, further tests show 
it is unnecessary to exploit deeper transformations. %

We conduct additional analysis of the proposed model, including the training strategy, running efficiency, the number of views,
the structure of the view set encoder and the effect of patch-level correlations, please refer to the supplementary material for more insights. 

\begin{table}
	\begin{center}
		\begin{tabular}{l c c}
			\toprule
			Decoder & Class Acc. (\%) & Inst. Acc. (\%) \\
			\midrule
			1-layer & 97.9 & 98.1 \\   %
			2-layer & \textbf{98.9} & \textbf{98.8} \\   %
			3-layer & \textcolor{blue}{98.5} & \textcolor{blue}{98.5} \\   %
			\bottomrule
		\end{tabular}
		\caption{Ablation study: choices for Decoder.}
		\label{tab:ablate_decoder}
	\end{center}
\end{table}

\subsection{Visualization}

\begin{figure}[ht]
	\begin{center}
		\includegraphics[width=0.9\linewidth]{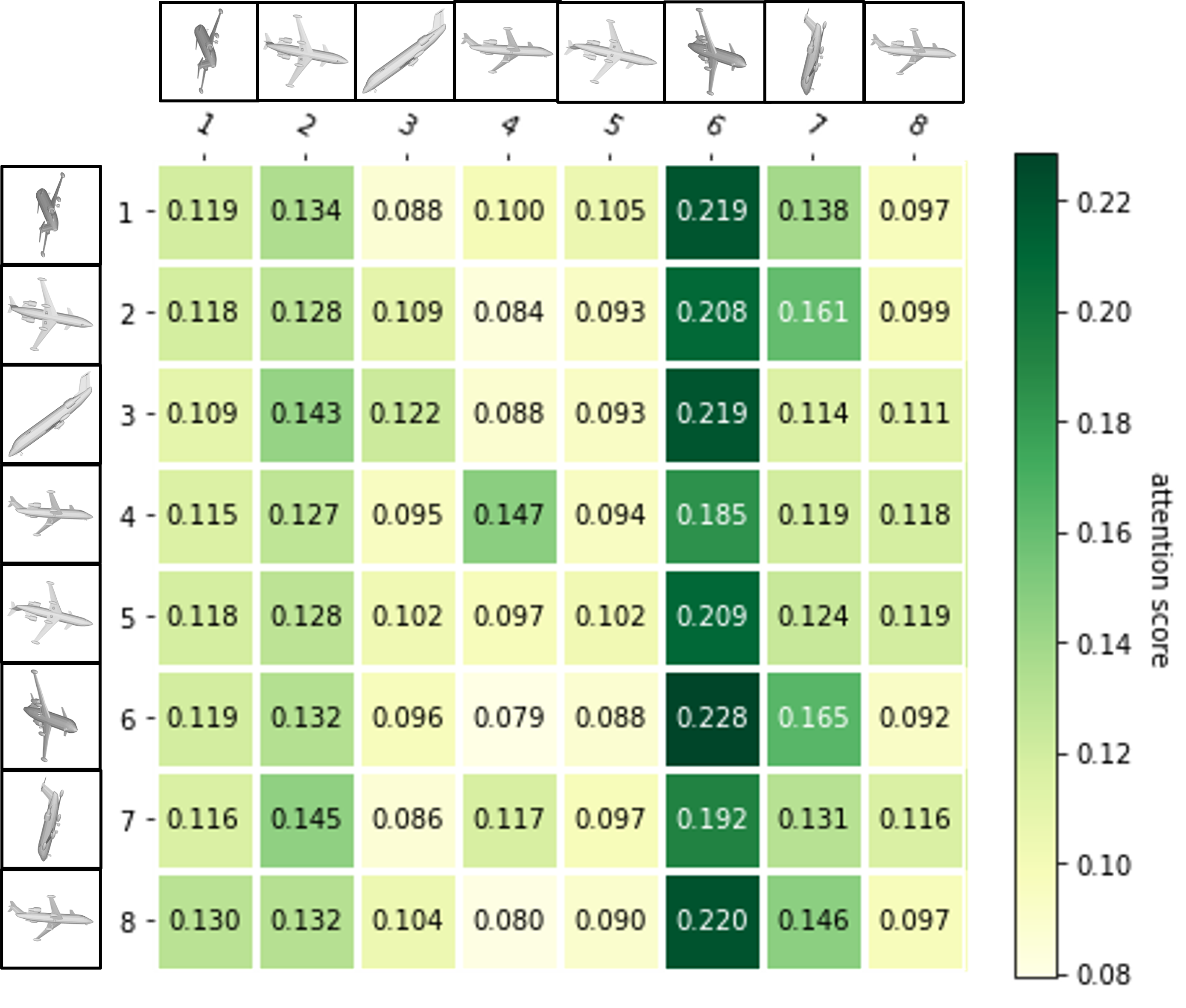}
	\end{center}
	\caption{Visualization of the attention scores for 8 views of a 3D airplane.}
	\label{fig:view_attn_score}
\end{figure}

\noindent\textbf{Multi-view Attention Map.} 
For better understanding, we visualize the attention map of eight views of a 3D airplane in Figure~\ref{fig:view_attn_score}. 
The attention scores are taken from the outputs of 
the last attention block of our model. The map indicates the 6th view is representative since it receives more attentions from other views. 
On the other hand, we can manually infer the 6th view is representative based on the visual appearances of these views. 
The results reflect that ViewFormer can adaptively capture the multi-view correlations and assign more weights to the representative views 
for recognition. 

\begin{figure}[t]
	\centering
	\begin{subfigure}{0.32\linewidth}
		\includegraphics[width=\linewidth]{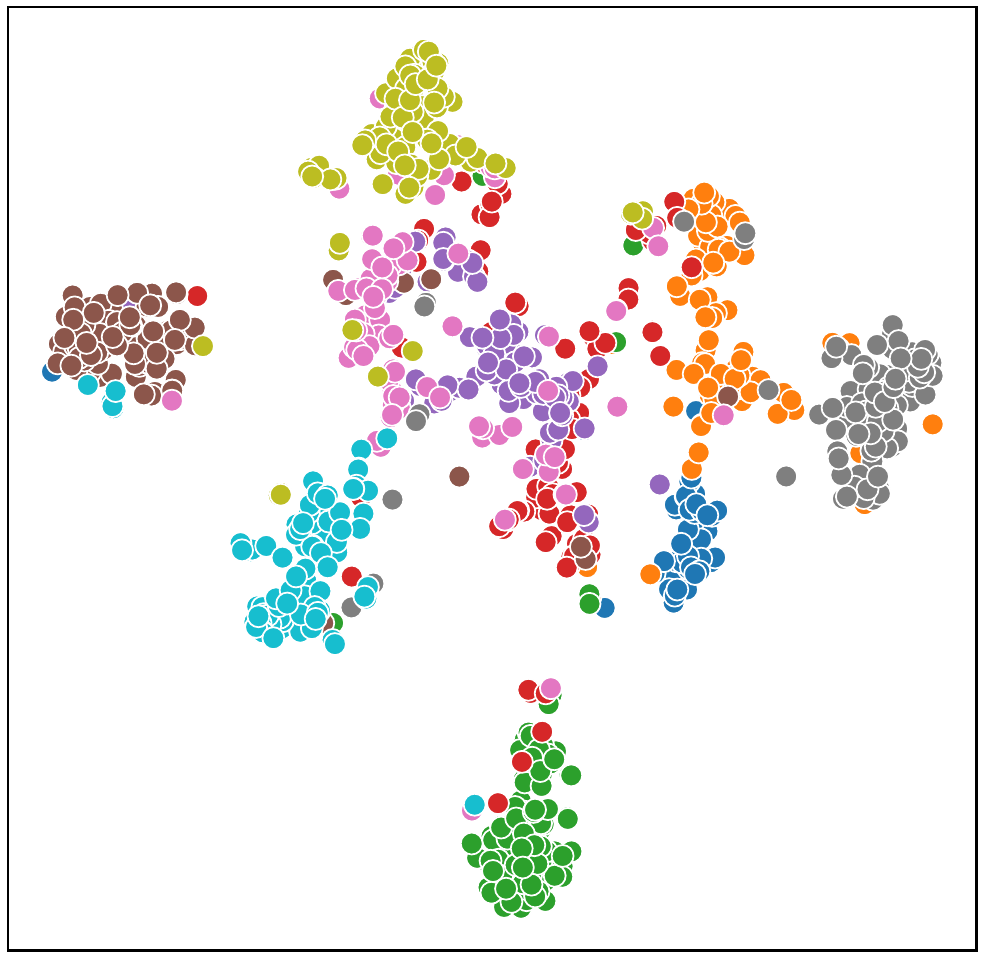}
		\caption{MN10}
		\label{fig:mn10_feats}
	\end{subfigure}
	\begin{subfigure}{0.32\linewidth}
		\includegraphics[width=\linewidth]{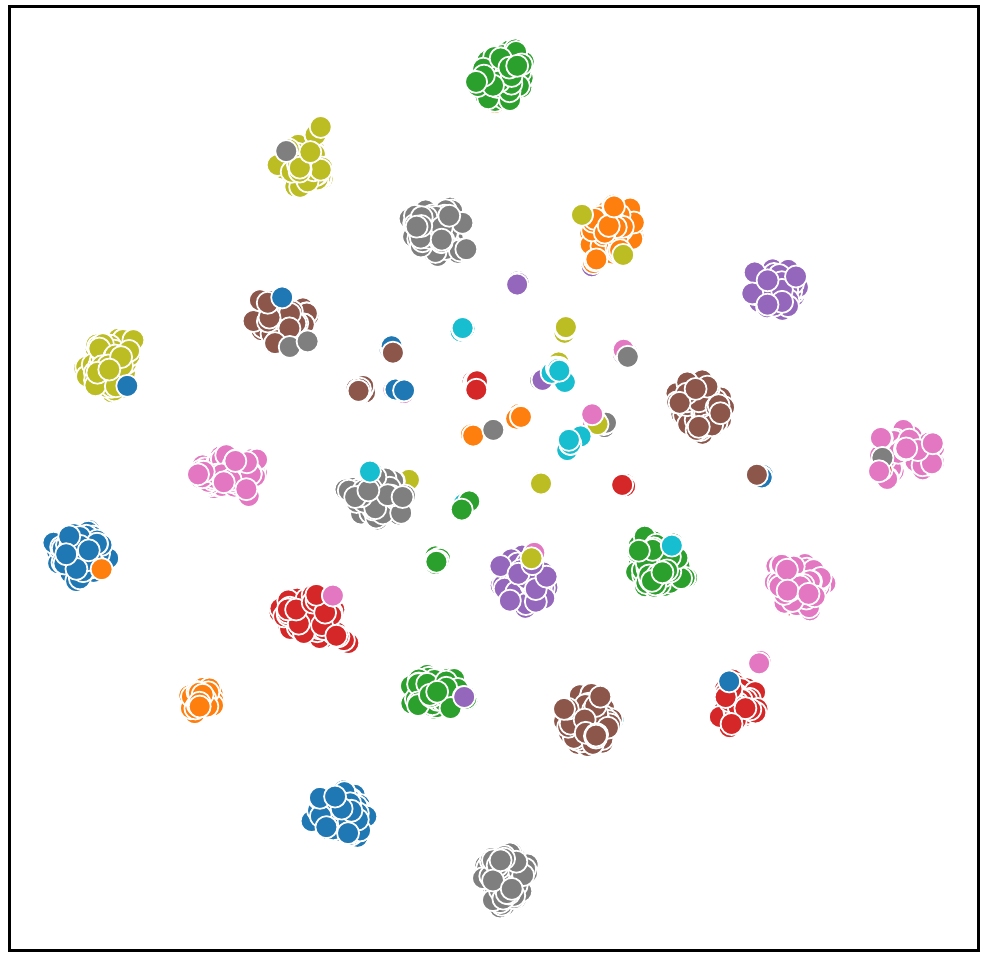}
		\caption{MN40}
		\label{fig:mn40_feats}
	\end{subfigure}
	\begin{subfigure}{0.32\linewidth}
		\includegraphics[width=\linewidth]{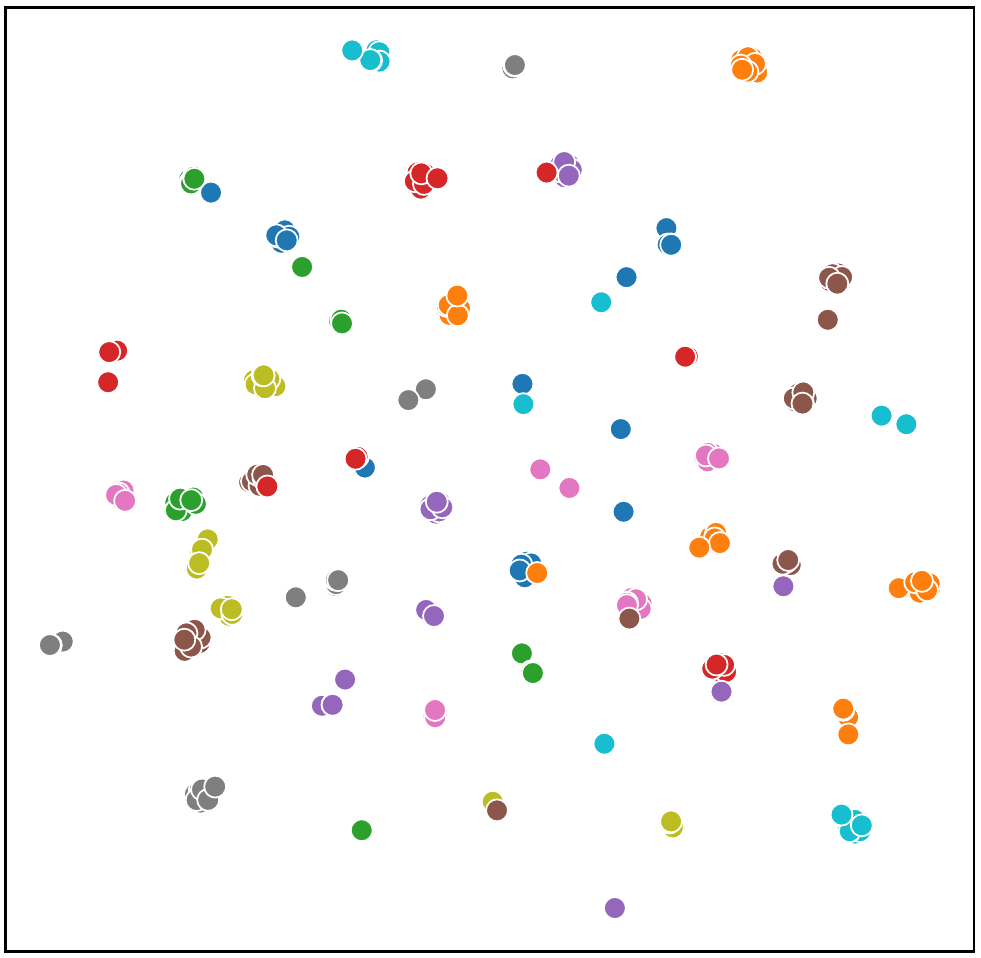}
		\caption{RGBD}
		\label{fig:rgbd_feats}
	\end{subfigure}
	\caption{Visualization of 3D shape feature distribution on ModelNet10 (MN10), ModelNet40 (MN40) and RGBD.}
	\label{fig:recog_feats_visual}
\end{figure}

\noindent\textbf{3D Shape Recognition.} 
We visualize the feature distribution for different shape categories on ModelNet10, ModelNet40 and RGBD using t-SNE~\cite{tsne}, 
shown in Figure~\ref{fig:recog_feats_visual}. 
It shows different shape categories of different datasets are successfully distinguished by the proposed method, 
demonstrating ViewFormer understands multi-view information well by 
explicitly modeling the correlations for all view pairs in the view set. 

\begin{figure}[t]
	\begin{center}
		\includegraphics[width=\linewidth]{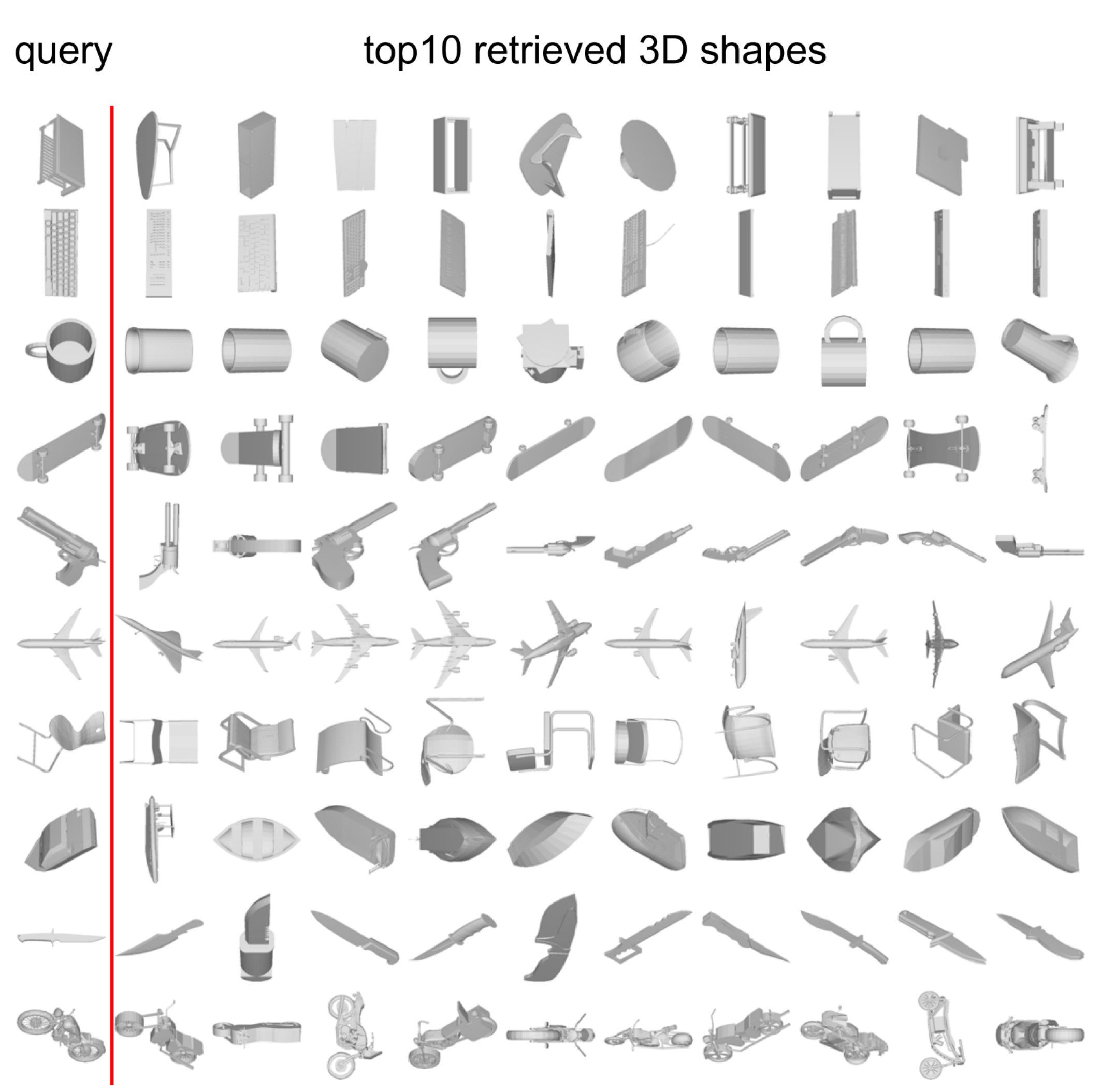}
	\end{center}
	\caption{Visualization of the top 10 retrieved results for each query shape.}
	\label{fig:shape_retrieval_example}
\end{figure}

\noindent\textbf{3D Shape Retrieval.} 
We visualize the top 10 retrieved shapes for 10 typical queries in Figure~\ref{fig:shape_retrieval_example}. 
The retrieval happens in the ShapeNet Core55 validation set. Each retrieved shape is represented by its random view. 
We find the top10 results are highly relevant to the query, which means they are in the same category. 
The 5th shape in the 3rd row maybe confusing, but actually, it is also a cup. Please refer to the supplementary 
material for more views of this shape. 

\section{Conclusion}
This paper presents ViewFormer, a simple yet effective multi-view 3D shape analysis method. A novel perspective is proposed 
to organize the multiple views of a 3D shape in a view set, which offers flexibility and avoids assumed relations for views.
Based on that, a view set attention model is devised to learn the pairwise and higher-order
correlations of the views in the set adaptively.
ViewFormer shows outstanding performances across different datasets and sets new records for recognition and retrieval tasks. 
But note that the performance gap between point/voxel-based and view-based methods is relatively large. 
In the future, we plan to explore cross-modal distillation between point/voxel-based and view-based models to narrow the gap. 

{\small
\bibliographystyle{ieee_fullname}
\bibliography{egbib}
}

\clearpage  %
\appendix
\section{Additional Analysis}
We provide additional analysis to the proposed approach, including network optimization, shallow convolution operations in Initializer, 
the number of views, learning efficiency, the architecture of the view set encoder, the performances gains delivered by the devised encoder, 
and the effect of patch-level feature interactions. 

\subsection{Network Optimization}
We adopt a 2-stage training strategy to optimize the proposed model and verify its effectiveness
through the following experiments. 

\begin{figure}[ht]
   \centering
   \includegraphics[width=\linewidth]{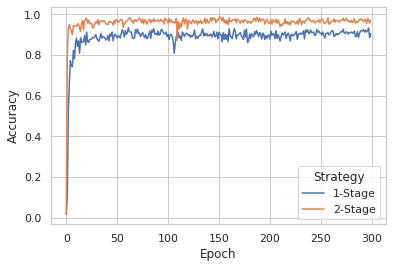}
   \caption{Comparison of instance accuracy on ModelNet40 when using 1-stage and 2-stage optimization.}
   \label{fig:viewformer_training_strategy}
\end{figure}

\noindent\textbf{1-Stage vs. 2-Stage.}
We compare the effectiveness of 1-stage and 2-stage optimization on ModelNet40. For 2-stage optimization, Initializer is trained 
individually on the dataset, then the pre-trained weights of Initializer are loaded into ViewFormer to be optimized with
other modules jointly. The 1-stage optimization means ViewFormer learns in an end-to-end way and all parameters are randomly initialized. 
Figure~\ref{fig:viewformer_training_strategy} shows the recognition accuracy achieved by 2-stage optimization is significantly 
better than that of 1-stage training. 
The results demonstrate ViewFormer receives gains from the well-initialized view representations 
provided by the first stage of learning.  

\begin{figure}[ht]
	\centering
   \includegraphics[width=\linewidth]{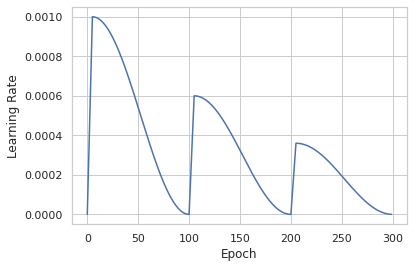}
   \caption{The learning rate curve for optimizing ViewFormer.}
   \label{fig:viewformer_lr}
\end{figure}

\noindent\textbf{Training Details.}
For Initializer, we train it 30 epochs on the target dataset using SGD~\cite{ruder2016overview}, with an initial learning rate 0.01 and 
CosineAnnealingLR scheduler. After that, the pre-trained weights of Initializer are loaded into ViewFormer to be optimized with other modules jointly. 
Specifically, ViewFormer is trained 300 epochs on the target dataset using AdamW~\cite{loshchilov2018decoupled}, 
with an initial peak learning rate 0.001 and CosAnnealingWarmupRestartsLR scheduler~\cite{Katsura21cawwr}. The restart interval is 100 epochs 
and the warmup happens in the first 5 epochs of each interval. 
The learning rate increases to the peak linearly during warmup and the peak decays by 40\% after each interval. 
The learning rate curve is visualized in Figure~\ref{fig:viewformer_lr}. 

\subsection{Shallow Convolution Operations in Initializer} 
We investigate the performances of ViewFormer when deploying shallow convolution operations as Initializer, \eg, 1- and 2-layer
convolution. Table~\ref{tab:init_1_layer_conv} and \ref{tab:init_2_layer_conv} explains their specific configurations. Due to 
the increased number of strides, 2-layer convolution has much lower parameters than 1-layer operation. However, ViewFormer with
shallow convolution initializations does not lead to decent 3D shape recognition. 
The best instance
accuracy is 93.7\%, much lower than 98.8\% given by ViewFormer with lightweight CNN (AlexNet) Initializer,
suggesting lightweight CNNs are reasonable choices for the Init module.

\begin{table}[ht]
   \begin{center}
      \begin{tabular}{l | l}
      \toprule
      View Size & $224\times224\times3$ \\
      \midrule
       \multirow{4}{*}{1st Conv} & Conv2d(in=3,out=64,k=7,s=2,p=3) \\ 
       & BatchNorm2d(num=64) \\
       & ReLU(inplace=True) \\   
       & MaxPool2d(k=3, s=2, p=1) \\  
      \midrule
      Total \#Params & 102.8 M \\
      \midrule
      Class Acc. & 90.1\% \\
      \midrule
      Inst. Acc. & 92.5\% \\
      \bottomrule
      \end{tabular}
   \caption{Configuration of the 1-layer convolution in Initializer.}  %
   \label{tab:init_1_layer_conv}
   \end{center}
\end{table}

\begin{table}[ht]
   \begin{center}
      \begin{tabular}{l | l}
      \toprule
      View Size & $224\times224\times3$ \\
      \midrule
       \multirow{4}{*}{1st Conv} & Conv2d(in=3, out=64, k=7, s=2, p=3) \\ 
       & BatchNorm2d(num=64) \\
       & ReLU(inplace=True) \\   
       & MaxPool2d(k=3, s=2, p=1) \\\hdashline
       \multirow{3}{*}{2nd Conv} & Conv2d(in=64, out=32, k=3, s=2, p=1) \\ 
       & BatchNorm2d(num=32) \\
       & ReLU(inplace=True) \\   
      \midrule
      Total \#Params & 12.9 M \\
      \midrule
      Class Acc. & 88.9\% \\
      \midrule
      Inst. Acc. & 93.7\% \\
      \bottomrule
      \end{tabular}
   \caption{Configuration of the 2-layer convolution in Initializer.}  %
   \label{tab:init_2_layer_conv}
   \end{center}
\end{table}

\subsection{Ablation Studies}

We conduct additional ablations to ViewFormer on ModelNet40, including the number of views used, 
the difference between various models using the same initializer module, 
the effect of pre-trained initializer and the performance gain brought by the encoder. 
We hope the studies can provide more insights on the design choices. 

\noindent\textbf{Effect of the Number of Views.}
We investigate the effect of the number of views on the recognition performance, shown in Table~\ref{tab:ablate_num_views}. 
There are up to 20 views for each 3D shape and we randomly select $M$ views for each shape for training and evaluation, 
where $M \in$ \{1, 4, 8, 12, 16, 20\}. 
When $M=1$, the problem is equivalent to single-view object recognition, so there is no interaction among views. In this case,
a lightweight ResNet18~\cite{he2016resnet} is trained for recognition and it achieves 89.0\% mean class accuracy and 91.8\% instance accuracy. 
When increasing the number of views, the performances are quickly improved. For instance, after aggregating the correlations from 4 views, 
ViewFormer lifts 8.4\% and 5.3\% absolute points in class and instance accuracy, respectively. 
But exploiting more views does not necessarily results in better accuracy. 
The 8-view ViewFormer reaches 98.0\% class accuracy and 98.8\% overall accuracy, outperforming 12- and 16-view versions. 
The performance is optimal when exploiting all 20 views and we choose this version to compare with other view-based methods. 

\begin{table}[ht]
   \begin{center}
      \begin{tabular}{c c c}
      \toprule
      \multirow{2}{*}{\#views} & Class Acc. & Inst. Acc. \\
      &  (\%) & (\%) \\
      \midrule
      1 & 89.0 & 91.8 \\   %
      4 & 97.4 & 97.1 \\
      8 & \textcolor{blue}{98.0} & \textbf{98.8} \\   
      12 & 97.5 & 97.6 \\   
      16 & 97.7 & \textcolor{blue}{98.3} \\
      20 & \textbf{98.9} & \textbf{98.8} \\
      \bottomrule
   \end{tabular}
   \caption{Ablation study: the number of views.}  %
   \label{tab:ablate_num_views}
   \end{center}
\end{table}

\noindent\textbf{Different Methods Using Same Initializer.} 
To be fair, we use same Initializer for different methods to inspect their recognition accuracies on ModelNet40. 
The chosen methods are representative baselines, RotationNet~\cite{Kanezaki18rotationnet} and View-GCN~\cite{wei20viewgcn}. 
The results in Table~\ref{tab:method_with_different_inits} show ViewFormer can achieve higher-level performance no matter 
the view representations are initialized by AlexNet~\cite{krizhevsky12alexnet} or ResNet50~\cite{he2016resnet}, 
exceeding View-GCN(AlexNet) and View-GCN(ResNet) by 1.6\% and 1.5\%, respectively. 
The results also indicate the proposed approach is better at grasping multi-view information for recognition since 
the initialized view features are identical. 

\begin{table}[ht]
   \begin{center}
      \begin{tabular}{l c c}
      \toprule
      Method & Initializer & Inst. Acc. (\%)\\
      \midrule
      RotationNet & \multirow{3}{*}{AlexNet} & 96.4 \\
      View-GCN &  & \textcolor{blue}{97.2} \\
      ViewFormer & & \textbf{98.8} \\\hdashline
      RotationNet & \multirow{3}{*}{ResNet50} & 96.9 \\
      View-GCN &  & \textcolor{blue}{97.3} \\
      ViewFormer & & \textbf{98.8} \\
      \bottomrule
      \end{tabular}
      \caption{Ablation study: different methods using a same Initializer.}  %
      \label{tab:method_with_different_inits}
   \end{center}
\end{table}

\begin{table*}[t]
   \centering
   \begin{tabular}{l r r r r r r r r r r r r r }
      \toprule
      \#Blocks & 2 & 2 & 2 & 2 & 4 & 4 & 4 & 4 & 6 & 6 & 6 & 6 \\ 
      \#Heads & 6 & 8 & 6 & 8 & 6 & 8 & 6 & 8 & 6 & 8 & 6 & 8 \\
      Ratio$_{mlp}$ & 2 & 2 & 4 & 4 & 2 & 2 & 4 & 4 & 2 & 2 & 4 & 4 \\
      Dim$_{view}$ & 384 & 512 & 384 & 512 & 384 & 512 & 384 & 512 & 384 & 512 & 384 & 512 \\
      \midrule
      \#Params (M) & 2.7 & 4.8 & 3.9 & 6.9 & 5.0 & 9.0 & 7.4 & 13.2 & 7.4 & 13.2 & 11.0 & 19.5\\
      \midrule
      ModelNet40 \\
      \quad Class Acc. (\%) & 98.8 & 98.7 & 98.4 & 97.2 & 97.4 & \textcolor{blue}{98.9} & \textbf{99.1} & 98.2 & 98.7 & 98.2 & 98.4 & 98.1 \\
      \quad Inst. Acc. (\%) & \textbf{99.0} & \textcolor{blue}{98.8} & 98.5 & 98.1 & 97.6 & \textcolor{blue}{98.8} & 98.5 & 98.5 & 98.3 & 98.3 & 98.1 & 98.3 \\
      \bottomrule
   \end{tabular}
   \caption{Ablation Study: the architecture of Encoder.}
   \label{tab:encoder_arch}
\end{table*}

\begin{figure}[ht]
	\centering
   \includegraphics[width=\linewidth]{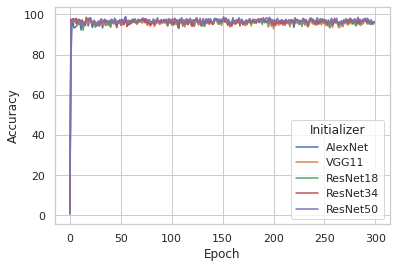}
   \caption{Learning efficiency of ViewFormer.}
   \label{fig:learning_efficiency}
\end{figure}

\noindent\textbf{Learning Efficiency.}
We explore the learning efficiency of ViewFormer by freezing the weights of the pre-trained Initializer. Figure~\ref{fig:learning_efficiency} 
displays the recognition accuracy curves of ViewFormer variants with different initializers on ModelNet40 during training. 
Regardless of Initializer used, all variants' performances soared after a short training and approached the highest. 
For instance, ViewFormer with ResNet34 Initializer reaches 97.6\% instance accuracy after only 2-epoch learning, while View-GCN~\cite{wei20viewgcn} achieves the
same performance with 7.5x longer optimization. The results reflect the proposed method has higher learning efficiency than the previous state of the art.  

\noindent\textbf{The Architecture of Encoder.} 
We provide ablations to justify the design choices of Encoder.
The controlled variables of Encoder are the number of attention blocks (\#Blocks), the number of attention heads in MSA (\#Heads), 
the widening ratio of MLP hidden layer (Ratio$_{mlp}$) and the dimension of the view representations (Dim$_{view}$). 
The mean class accuracy and instance accuracy of ViewFormer with different encoder structures are compared in Table~\ref{tab:encoder_arch}. 
All design variants show high-level performances and surpass the existing state of the art. 
Surprisingly, the encoder consisting of only 2 attention blocks can facilitate ViewFormer to achieve 99.0\% overall accuracy. 
The results are in line with expectations as the size of a view set is relatively small thus, it is unnecessary to 
design a very complex encoder. At the same time, it is inspiring that pairwise and higher-order correlations of elements in the view set
can be enriched and well grasped by a shallow encoder. Finally, we select the design that takes \emph{the second place} in both mean class and 
instance accuracy, namely \#Blocks = 4, \#Heads = 8, Ratio$_{mlp}$ = 2 and Dim$_{view}$ = 512. 

\noindent\textbf{Performance Gains Delivered by Our Encoder.} 
We investigate the performance gains delivered by the devised view set encoder. First, the initializer is individually trained to recognize 3D
shapes in ModelNet40. Second, the devised encoder is appended upon the pre-trained initializer to 
further capture the feature interactions among views. The chosen architecture for encoder is 
\#Blocks = 2, \#Heads = 6, Ratio$_{mlp}$ = 2, Dim$_{view}$ = 384, seen in Table~\ref{tab:encoder_arch}. 
Table~\ref{tab:performance_gains_by_encoder} compares the number of parameters and performances of different configurations described above.
Notable performance gains are obtained by the proposed view set encoder over different initializers. 
For example, by appending 2 attention blocks on the AlexNet initializer, our model achieves 18.2\% and 14.9\% absolute improvements for mean class accuracy 
and instance accuracy.
In contrast, the introduced 2.7M parameters only account for 6.4\% of that in AlexNet~\cite{krizhevsky12alexnet}. 

\begin{table}[ht]
   \begin{center}
      \begin{tabular}{l c c c}
      \toprule
      \multirow{2}{*}{Module} & \#Params & Class Acc. & Inst. Acc. \\
      & (M) & (\%) & (\%)\\
      \midrule
      AlexNet & \textcolor{blue}{42.3} & \textcolor{blue}{80.6} & \textcolor{blue}{85.1} \\  %
      $~$+ 2 Attn. Blocks & \textbf{2.7} & \textbf{98.8} & \textbf{99.0} \\   %
      \midrule
      ResNet18 & \textcolor{blue}{11.2} & \textcolor{blue}{88.7} & \textcolor{blue}{91.8} \\ %
      $~$+ 2 Attn. Blocks & \textbf{2.7} & \textbf{98.1} & \textbf{97.8} \\  %
      \bottomrule
   \end{tabular}
   \caption{Ablation study: the performance gains brought by the devised encoder over Initializer.}  %
   \label{tab:performance_gains_by_encoder}
   \end{center}
\end{table}

\noindent\textbf{Effect of Patch-level Feature Correlations.}
Some other methods also consider fine-grained patch-level interactions~\cite{yu18mhbn,yu21mhbn++,chen21mvt,xu21carnet}. They believe 
multi-view information flow can be enhanced by integrating patch-level features. In this work, we examine the effect of 
patch-level feature correlations by injecting them into each attention block of the encoder. The results in 
Table~\ref{tab:patch_level_correlations} show injecting patch-level features is redundant and unnecessary, 
disturbs the multi-view information understanding and decreases the performance slightly. But whether the patch-level correlations are 
integrated or not, ViewFormer maintains high-level performances (98.0+\% accuracies) and surpasses all existing models. 

\begin{table}
   \begin{center}
      \begin{tabular}{l c c}
      \toprule
      Variants & Class Acc. (\%) & Inst. Acc. (\%)\\
      \midrule
      w/ patch & \textcolor{blue}{98.1} & \textcolor{blue}{98.1}\\    %
      w/o patch & \textbf{98.9} & \textbf{98.8}\\ %
      \bottomrule
   \end{tabular}
   \caption{Ablation study: effect of the patch-level correlations.}
   \label{tab:patch_level_correlations}
   \end{center}
\end{table}

\section{Visualizations}

\begin{figure}[ht]
	\centering
   \includegraphics[width=\linewidth]{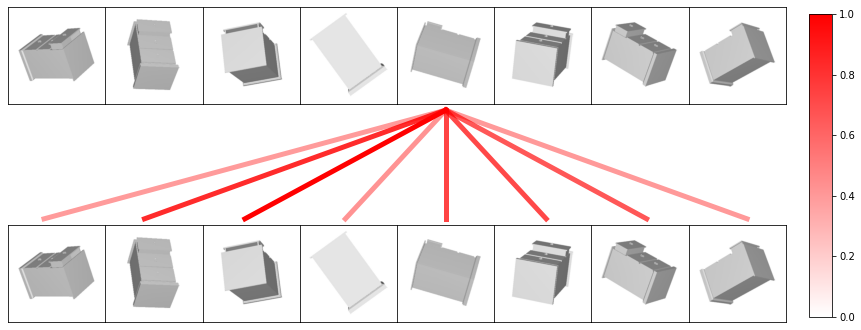}
   \caption{Visualization of multi-view attention of 8 views of a nightstand in colored lines.}
   \label{fig:8nightstands_attn_lines}
\end{figure}

\noindent\textbf{Multi-view Attention in Colored Lines.}
We randomly select a 3D shape that is a nightstand, then visualize the multi-view correlations of eight views of this 
shape, referring to Figure~\ref{fig:8nightstands_attn_lines}. The correlations are represented by the attention scores emitted 
by the last attention block of ViewFormer. The scores are normalized to map to the color bar on the far right of the figure. 
Our model distributes more weights to 2nd, 3rd, 6th views from the 5th one. The results seem reasonable since these views 
are more discriminative according to visual appearances.

\begin{figure}[ht]
	\centering
   \includegraphics[width=\linewidth]{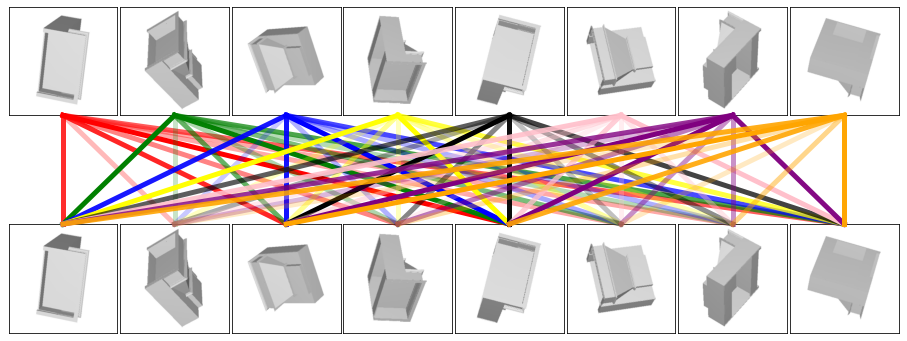}
   \caption{Visualization of multi-view attention for all view pairs of a range hood in colorful lines.}
   \label{fig:8rangehoods_attn_all_lines}
\end{figure}

Another 3D shape is randomly selected to demonstrate multi-view attention. The selected shape is a range hood. 
In Figure~\ref{fig:8rangehoods_attn_all_lines}, we visualize the interactions of all view pairs for the shape. 
The purpose is to let readers feel the flexibility of organizing multiple views of a 3D shape into a set and the 
powerful capability of view set attention in modeling the correlations among elements in a set. 

\begin{figure}[ht]
	\centering
	\begin{subfigure}{0.19\linewidth}
		\includegraphics[width=\linewidth]{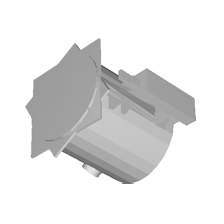}
	\end{subfigure}
	\begin{subfigure}{0.19\linewidth}
		\includegraphics[width=\linewidth]{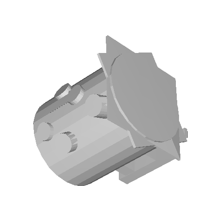}
	\end{subfigure}
	\begin{subfigure}{0.19\linewidth}
		\includegraphics[width=\linewidth]{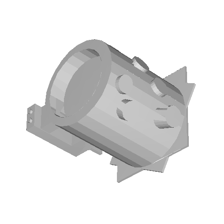}
	\end{subfigure}
   \begin{subfigure}{0.19\linewidth}
		\includegraphics[width=\linewidth]{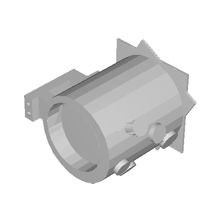}
	\end{subfigure}
	\begin{subfigure}{0.19\linewidth}
		\includegraphics[width=\linewidth]{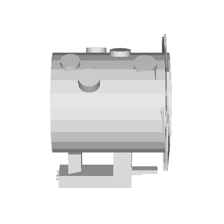}
	\end{subfigure}

	\begin{subfigure}{0.19\linewidth}
		\includegraphics[width=\linewidth]{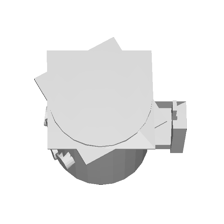}
	\end{subfigure}
   \begin{subfigure}{0.19\linewidth}
		\includegraphics[width=\linewidth]{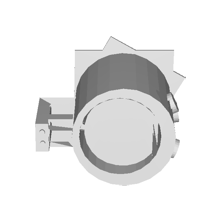}
	\end{subfigure}
	\begin{subfigure}{0.19\linewidth}
		\includegraphics[width=\linewidth]{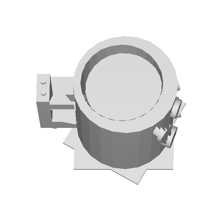}
	\end{subfigure}
	\begin{subfigure}{0.19\linewidth}
		\includegraphics[width=\linewidth]{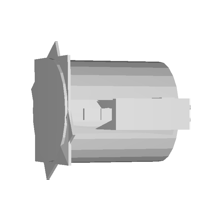}
	\end{subfigure}
   \begin{subfigure}{0.19\linewidth}
		\includegraphics[width=\linewidth]{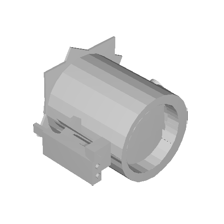}
	\end{subfigure}

	\caption{Multiple views of a retrieved shape.}
	\label{fig:views_of_a_cup}
\end{figure}

\noindent\textbf{Multiple Views of a Retrieved Shape.}
In Figure \textcolor{red}{4} of the main paper, the retrieved shape in the 5th column of the 3rd row may be confusing since 
one may not be able to determine whether it belongs to the same class as the query. 
To this end, we pinpoint the shape in the dataset and find more views of it, shown in Figure~\ref{fig:views_of_a_cup}. 
After observing these views, we can infer this shape is a cup, so it is of the same class as the query. 
The example also demonstrates a central problem of multi-view 3D shape analysis, how to integrate multi-view information effectively. 

\end{document}